%% file: main.tex
\documentclass[conference]{IEEEtran}
\IEEEoverridecommandlockouts
\usepackage{cite}
\usepackage{amsmath,amssymb,amsfonts}
\usepackage{graphicx}
\usepackage{textcomp}
\usepackage{xcolor}
\usepackage{float}
\usepackage{booktabs}
\usepackage{tabularray,varwidth,enumitem}
\usepackage{multirow, makecell}
\usepackage{algorithm}
\usepackage{algpseudocode}
\usepackage{subcaption}
\usepackage{url}
\usepackage{soul}
\usepackage[switch]{lineno} 

\usepackage{caption} 

\def\BibTeX{{\rm B\kern-.05em{\sc i\kern-.025em b}\kern-.08em
    T\kern-.1667em\lower.7ex\hbox{E}\kern-.125emX}}

\begin{document}
\bstctlcite{IEEEexample:BSTcontrol} 
\nolinenumbers

\title{FedCert: Federated Accuracy Certification}

\newcommand{\lre}[2]{\st{#1}\textcolor{blue}{#2}}
\newcommand{\lde}[1]{\textcolor{blue}{\st{#1}}}
\newcommand{\ladd}[1]{\textcolor{blue}{#1}}
\newcommand{\lcom}[1]{\textcolor{red}{Le's comment: #1}}
\newcommand{\done}[1]{\textcolor{cyan}{#1}}
\newcommand{\nguyen}[1]{\textcolor{green}{#1}}

\author{
\IEEEauthorblockN{
Minh Hieu Nguyen\IEEEauthorrefmark{1}\IEEEauthorrefmark{4},
Huu Tien Nguyen\IEEEauthorrefmark{1}\IEEEauthorrefmark{4},
Trung Thanh Nguyen\IEEEauthorrefmark{2},
Manh Duong Nguyen\IEEEauthorrefmark{1},\\
Trong Nghia Hoang\IEEEauthorrefmark{5},
Truong Thao Nguyen\IEEEauthorrefmark{6}\IEEEauthorrefmark{8},
Phi Le Nguyen\IEEEauthorrefmark{1}\IEEEauthorrefmark{8} \thanks{\IEEEauthorrefmark{8}Corresponding author.}
}
\IEEEauthorblockA{
\IEEEauthorrefmark{1} Hanoi University of Science and Technology, Hanoi, Vietnam
\\\{hieu.nm194049, tien.nh205033, duong.nm210243\}@sis.hust.edu.vn, lenp@soict.hust.edu.vn}
\IEEEauthorblockA{\IEEEauthorrefmark{2} Nagoya University, Nagoya, Japan, nguyent@cs.is.i.nagoya-u.ac.jp}
\IEEEauthorblockA{\IEEEauthorrefmark{5} Washington State University, State of Washington, United States, trongnghia.hoang@wsu.edu}
\IEEEauthorblockA{\IEEEauthorrefmark{6} National Institute of Advanced Industrial Science and Technology (AIST), Japan, nguyen.truong@aist.go.jp}
}

\maketitle
\thispagestyle{plain}
\pagestyle{plain}
\begingroup\renewcommand\thefootnote{\textsection}
\footnotetext{The first and second authors contributed equally to this research.}

\endgroup

\begin{abstract}
Federated Learning (FL) has emerged as a powerful paradigm for training machine learning models in a decentralized manner, preserving data privacy by keeping local data on clients. 
However, evaluating the robustness of these models against data perturbations on clients remains a significant challenge. 
Previous studies have assessed the effectiveness of models in centralized training based on certified accuracy, which guarantees that a certain percentage of the model's predictions will remain correct even if the input data is perturbed. 
However, the challenge of extending these evaluations to FL remains unresolved due to the unknown client's local data.
To tackle this challenge, this study proposed a method named FedCert to take the \textit{first} step toward evaluating the robustness of FL systems. 
The proposed method is designed to approximate the certified accuracy of a global model based on the certified accuracy and class distribution of each client. 
Additionally, considering the Non-Independent and Identically Distributed (Non-IID) nature of data in real-world scenarios, we introduce the client grouping algorithm to ensure reliable certified accuracy during the aggregation step of the approximation algorithm.
Through theoretical analysis, we demonstrate the effectiveness of FedCert in assessing the robustness and reliability of FL systems. Moreover, experimental results on the CIFAR-10 and CIFAR-100 datasets under various scenarios show that FedCert consistently reduces the estimation error compared to baseline methods.
This study offers a solution for evaluating the robustness of FL systems and lays the groundwork for future research to enhance the dependability of decentralized learning.
\end{abstract}


\section{Introduction}
\label{sec:introduction}
\input{Chapters/1-Introduction}

\section{Background and Related Work}
\label{sec:preliminaries}
\input{Chapters/2-Preliminaries}

\section{Methodology}
\label{sec:mothodology}
\input{Chapters/3-1-Theorem}

\input{Chapters/3-2-Methodology}

\section{Evaluation}
\label{sec:evaluation}
\input{Chapters/4-Experiments}

\section{Conclusion}
\label{sec:conclusion}
\input{Chapters/6-Conclusion}

\section*{Acknowledgment}
This research is funded by Hanoi University of Science and Technology (HUST) under grant number T2023-PC-028.
This work was funded by Vingroup Joint Stock Company (Vingroup JSC),Vingroup, and supported by Vingroup Innovation Foundation (VINIF) under project code VINIF.2021.DA00128.

\bibliographystyle{IEEEtran}
\bibliography{references}

\end{document}

%% file: Chapters/1-Introduction.tex
In recent years, Federated Learning (FL)~\cite{mcmahan2017communication} has emerged as a promising privacy-preserving learning paradigm that enables multiple clients to collaboratively train Machine Learning (ML) models without sharing their data. 
FL is particularly advantageous in scenarios where data privacy is a significant concern, such as in healthcare~\cite{rahman2023federated, antunes2022federated} and finance~\cite{long2020federated, imteaj2022leveraging}. 
As a result, it has been widely applied in various applications due to its efficiency and privacy-preserving properties.
Despite its advantages, FL faces challenges in accurately evaluating the system's robustness.
This difficulty arises primarily from the vulnerability of ML models used by the global server and clients in FL to adversarial attacks. 
This issue stems from that ML models can produce vastly different predictions for inputs that the human eye can not distinguish due to small adversarial perturbations~\cite{goodfellow2014explaining}.
To ensure the robustness of ML models, certified accuracy~\cite{lecuyer2019certified, li2019certified} is a concept used to measure their robustness.
Certified accuracy guarantees that a certain percentage of the model’s predictions will remain correct even if the input data is perturbed within a specified radius. 

For FL systems, the Volume-based Weighted-sum (VW) method~\cite{https://doi.org/10.48550/arxiv.2210.13291} is a potential approach for evaluating the certified accuracy of a global model. This method combines the certified accuracy of individual clients, weighted by the size of their respective test datasets. 
While this approach effectively preserves the privacy of each client's data, it faces significant challenges in generalizability. 
Specifically, VW struggles to maintain accuracy and reliability when client data is highly heterogeneous, a scenario frequently encountered in real-world FL applications. 
In such situations, VW leads to less reliable evaluations of the global model's performance.
To address these limitations, we propose FedCert, which more accurately evaluates the global model's certified accuracy by solving convex optimization problems. 
The proposed method considers each client's certified accuracy and class distribution to provide a comprehensive and reliable assessment. 
Additionally, to address the Non-IID nature of data in real-world scenarios, we introduce a client grouping algorithm that mitigates variability in client data and enhances the reliability of the evaluation process. 
By addressing these challenges, we aim to improve the robustness and reliability of FL systems against adversarial threats, providing a foundation for more secure and trustworthy FL applications.

In this study, we take the \textit{first} step towards evaluating certified accuracy in FL. 
We make contributions on both theoretical and experimental fronts, as follows:
\begin{itemize}
    \item We perform a theoretical analysis of the current limitations in evaluating certified accuracy in FL and provide proof for our motivation.
    \item We propose a method named FedCert to evaluate the certified accuracy of FL systems. 
    The proposed method approximates the certified accuracy of a global model based on the certified accuracy and class distribution of each client. 
    Additionally, we introduce a client grouping method to ensure reliable certified accuracy during the aggregation step of the approximation algorithm.
    \item We conduct comprehensive experiments on the well-known CIFAR-10 and CIFAR-100 datasets across various scenarios to evaluate the effectiveness of FedCert.
\end{itemize}

%% file: Chapters/2-Preliminaries.tex
\subsection{Federated Learning}
Federated Learning (FL)~\cite{mcmahan2017communication} is a training paradigm that allows multiple clients (data holders) to collaboratively train a model in a distributed manner while preserving data privacy. 
Unlike traditional centralized training methods, FL enables clients to train models using local data and only share model parameters with a central server. 
The training process in FL involves multiple communication rounds, each consisting of a \textit{local training} phase at the client side and an \textit{aggregation} phase at the server side.
At the start of each round $t$, every client receives the global model $\theta^t$ from the server and applies a learning algorithm (such as gradient descent) to update the model using its data.
After completing the local training step, each client $C_i$ obtains its local model $\theta_i^t$ and sends it to the server for the aggregation step. 
The simplest aggregation method involves weighted averaging~\cite{mcmahan2017communication}, where each client's contribution is proportional to the size of its data, as follows:
\begin{equation}
\nonumber
    \theta^{t+1} = \sum_{i=1}^{N}\frac{|D_i|}{\sum_{i=j}^{K}|D_j|} \theta^{t}_i,
\end{equation}
where $N$ is the number of clients, $D_i$ denotes the dataset of client $C_i$, and $|.|$ represents the cardinality.

\subsection{Certified Accuracy}
Deep learning models are inherently susceptible to perturbations in their input data. 
Conventional performance metrics (e.g., prediction accuracy) fail to fully capture a model's effectiveness in real-world scenarios where noise is ubiquitous. 
To address this limitation, the concept of \textit{``certified accuracy''} has been introduced as a robust metric for evaluating a model's generalizability under input perturbations~\cite{li2019certified, lecuyer2019certified}. 
The certified accuracy of a classifier \( f \) at a test radius \( r \), which represents the maximum allowable perturbation in the input data,  is denoted as \( c(f, S, r) \) for a given dataset \( S \). 
It is defined as the proportion of \( S \) for which \( f \) is provably robust within an \( l_2 \) ball of radius \( r \)\footnote{A prediction is considered provably robust within an \( l_2 \) ball of radius \( r \) if the classifier \( f \) maintains accurate predictions even when the input is perturbed by random noise \( \varepsilon \sim \mathcal{N}(0, rI)\),  where  $I$ is the identity matrix.}. 
Formally, the certified accuracy \( c(f, S, r) \) is expressed as:
\begin{equation}
\nonumber
    c(f, S, r) = \frac{n_S^{\text{robust}}}{n_S},
\end{equation}
where \( n_S^{\text{robust}} \) denotes the number of samples for which \( f \) provides correct and certifiably robust predictions within an \( l_2 \) ball of radius \( r \), and \( n_S \) is the number of samples in \( S \).
\subsection{Robustness in Federated Learning}
Research on the robustness of FL primarily examines the effects of adversarial attacks on the FL system and devises algorithms to address them. 
One approach is Federated Adversarial Training (FAT)~\cite{li2023federated}, which deploys an adversarial training scheme on local clients for the conventional FL algorithm FedAvg~\cite{mcmahan2017communication}.
Moreover, Chen et al.~\cite{chen2021certifiablyrobust} integrated randomized smoothing into FL and applied adversarial training to update the local model.
This approach uses a volume-based aggregated global model similar to FedAvg to evaluate the robustness of the FL system.
Additionally, Alfarra et al.~\cite{alfarra2022certified} investigated the benefits of FL on certified robustness using diverse perturbation methods, including Gaussian noise, rotation, and pixel perturbations locally. 
Several studies have also theoretically analyzed the robustness of FL under noise. 
Yin et al.\cite{yin2018byzantine} developed a robust distributed optimization algorithm against arbitrary adversarial behavior and focused on achieving optimal statistical performance. 
Reisizadeh et al.~\cite{reisizadeh2020robust} introduced a robust FL algorithm by considering the structured affine distribution shift in users' data.

%% file: Chapters/3-1-Theorem.tex
\noindent \textbf{Problem Definition.} 
Consider a Federated Learning (FL) system with \( N \) clients, where each client \( i \) has a local dataset \( D_i \). This study focuses on the classification problem, specifically in scenarios where the datasets \( D_i \) (for \( i = \{1, \ldots, N\} \)) contain the same set of classes. Let \( \theta \) represent the global model obtained after the FL process. Our goal is to certify the accuracy of \( \theta \) when it is deployed in practice. However, since the server does not have access to a test dataset, conventional accuracy certification methods cannot be applied. To this end, we assume the server has knowledge of the class distribution in practice, denoted as \( p(S) \), where \( S \) represents the data in practice (this assumption is reasonable, as class distributions often follow a uniform distribution). Given \( p(S) \), our task is to estimate the certified accuracy of \( \theta \) with respect to \( S \) within a specified test radius \( r \).

\noindent \textbf{Motivation.} As previously mentioned, directly certifying the global model on the dataset \( S \) is not feasible because, in a FL context, the server typically does not have access to the data. To this end, our main approach involves allowing each client to certify the accuracy of $\theta$ on its local dataset. These local certified accuracy are then combined linearly by the server to produce the estimated certified accuracy of $\theta$ on \( S \).
Specifically, let \( c(\theta, D_i, r) \) represents the accuracy of $\theta$ certified by client \( i \) on its local dataset \( D_i \) within radius \( r \), the certified accuracy of $\theta$ with respect to \( S \) can then be estimated using the following formula:
\begin{equation}
    \nonumber
    c(\theta, S, r) \approx \sum_{i=1}^N{\alpha_i}c(\theta,D_i, r).
\end{equation}
Our problem is reduced to finding the values of \(\alpha_i\) (\(i = 1, \dots, N\)) satisfying the following objective function:
\begin{equation}
    \nonumber
    \{\alpha^*_i\}_{i=1}^{N} = \underset{\{\alpha_i\}_{i=1}^{N}}{\text{argmin}}{\left\|c(\theta, S, r) - \sum_{i=1}^{N}\alpha_ic(\theta, D_i, r) \right\| }.
\end{equation}
The key idea behind our solution for determining optimal values of $\alpha_i$ is rooted in analyzing the class distribution of each client's local dataset.
In the following sections, we first describe our theoretical analysis and then present the details of our practical algorithm.

\subsection{Theoretical Analysis}
\label{subsec:theoretical_analysis}
In this section, we provide our observations and theoretical analysis concerning the properties of \(\alpha_i\), which form the basis for our design of \(\alpha_i\) determination algorithm.

\noindent \textbf{Lemma 1.} 
Let $(D_1, \ldots, D_N)$ be arbitrary datasets, and $D$ be their union, i.e., $D = \{D_1 \cup D_2 \cup \ldots \cup D_N\}$.
Then, the certified accuracy of an arbitrary model $\theta$ on $D$ is a linear combination of those on $(D_1, \ldots, D_N)$.

\noindent \textbf{Proof.}
Let $n_i$ be the cardinality of $D_i$ $(i = \{1, \dots, N\})$, and $n$ be the cardinality of $D$, then the following holds:
\begin{equation}
    \nonumber
    c(\theta, D, r) = \sum_{i=1}^N{\frac{n_i}{n}c(\theta, D_i, r)}.
\end{equation}

\noindent \textbf{Lemma 2.} 
Let  \( \left(p(D_1), \dots, p(D_N)\right) \) be $N$ class distributions and \( p(S) \) be another class distribution which can be represented as a linear combination of $p(D_i)$ as follows:
\begin{equation}
    \nonumber
    p(S) = \sum_{i=1}^{N}{\alpha_i p(D_i)}, \quad \sum_{i=1}^{N}{\alpha_i} = 1, \quad 0 \leq \alpha_i \leq 1.
\end{equation}
The relationship between the certified accuracy of a model \(\theta\) on the dataset \(S\) with distribution \(p(S)\) and its certified accuracy on the datasets \((D_1, \dots, D_N)\) with distributions \((p(D_1), \dots, p(D_N))\) is given as follows:
\begin{equation}
    \nonumber
    \mathbf{E}_{S \sim p(S)}[c(\theta,S, r)] = \sum_{i=1}^{N}{\alpha_i \mathbf{E}_{D_i \sim p(D_i)}[c(\theta,D_i, r)]}.
\end{equation}

\noindent \textbf{Proof.} 
Let \( S^j \) be the set of samples in S belonging to class \( j \) for \( j = \{1, \ldots, M \}\), where \( M \) is the number of classes. 
Let \( a_j(S) \) denote the proportion of \( S^j \) in \( S \).
From Lemma 1, we have: $c(\theta,S,r) = a_j(S)\sum_{j=1}^M c(\theta, S^j, r)$, therefore, by the linearity of expectation, we have:
\begin{equation}
\label{eq:lemma_2_1}
    \mathbf{E}_{S \sim p(S)}[c(\theta,S, r)] = \sum_{j=1}^M a_j(S) \mathbf{E}_{S^j \sim p(S^j)}[c(\theta,S^j, r)],
\end{equation}
\begin{equation}
\label{eq:lemma_2_2}
    \mathbf{E}_{D_i \sim p(D_i)}[c(\theta,D_i, r)] = \sum_{j=1}^M a_j(D_i) \mathbf{E}_{D_i^j \sim p(D_i^j)}[c(\theta,D_i^j, r)].
\end{equation}
Because $p(S^j) = p(D^j_i)$, we have:
\begin{equation}
\label{eq:lemma_2_3}
    \mathbf{E}_{S^j \sim p(S^j)}[c(\theta,S^j, r)] = \mathbf{E}_{D_i^j \sim p(D_i^j)}[c(\theta,D_i^j, r)].
\end{equation}
And we have \( p(S) = \sum_{i=1}^{N} \alpha_i p(D_i) \), which means:
\begin{equation}
\label{eq:lemma_2_4}
    a_j(S) = \sum_{i=1}^{N} \alpha_i a_j(D_i).
\end{equation}
By substituting \eqref{eq:lemma_2_3} and \eqref{eq:lemma_2_4} into \eqref{eq:lemma_2_1}, we obtain:

\begin{equation}
\label{eq:lemma_2_5}
\begin{split}
    \mathbf{E}_{S \sim p(S)}[c(\theta,S, r)] 
    = \sum_{i=1}^N \alpha_i \sum_{j=1}^M a_j(D_i)  \mathbf{E}_{D_i^j \sim p(D_i^j)}[c(\theta,D_i^j, r)].
\end{split}
\end{equation}
From Eq.~\eqref{eq:lemma_2_2} and Eq.~\eqref{eq:lemma_2_5}, Lemma 2 is proved.


Next, we present a theorem that establishes a bound on the estimation error.


\noindent \textbf{Theorem 1.} 
Let  \((p(D_1), \dots, p(D_N))\) be $N$ class distributions and \( p(S) \) be another class distribution. 
Let \((\alpha_1, \dots, \alpha_N)\) be arbitrary numbers, and let \(\delta\) represent the difference between \(p(S)\) and the linear combination of \((p(D_1), \dots, p(D_N))\), defined as \(\delta = || p(S) - \sum_{i=1}^N \alpha^*_i p(D_i) || \). 
Then, there exists a constant \(Q\), independent of \((\alpha_1, \dots, \alpha_N)\), that satisfies the following inequality:
\begin{equation}
\label{eq:inequation}
    \left\| \mathbf{E}_{S \sim p(S)}[c(\theta,S, r)] - \sum_{i=1}^N \alpha^*_i \mathbf{E}_{D_i \sim p(D_i)}[c(\theta,D_i, r)] \right\| \leq \delta Q.
\end{equation}



\noindent \textbf{Proof.}
Let \( S' \) be a dataset whose class distribution, $p(S')$, is a linear combination of \((p(D_1), \dots, p(D_N))\), such that \( p(S') = \sum_{i=1}^N \alpha^*_i p(D_i) \). Let us denote by \( p_j \) the probability of class \( j \) under the distribution \( p(S) \), and \( p'_j \) the probability of class \( j \) under the distribution \( p(S') \) ( $j = 1, ..., M$, where \( M \) is the total number of classes). Additionally, let \( \mathbb{M}_j \) be an arbitrary dataset consisting only samples with label \( j \).
To easy the presentation, we denote $L_j(r) = \mathbf{E}_{\mathbb{M}_j }[c(\theta, \mathbb{M}_j, r)]$.
Applying Lemma 2, we obtain:
\begin{eqnarray}
\nonumber
        \mathbf{E}_{S \sim p(S)}[c(\theta,S, r)] & = & \sum_{j=1}^{M} p_j L_j(r), \\
        \mathbf{E}_{S' \sim p(S')}[c(\theta, S', r)] & = & \sum_{j=1}^{M} p'_{j} L_j(r).
\end{eqnarray}
The left side of (\ref{eq:inequation}) then can be represented as follows:
\begin{align*}
    H &= \left \|\mathbf{E}_{S \sim p(S)}[c(\theta,S, r)] - \mathbf{E}_{S' \sim p(S')}[c(\theta, S', r)] \right \| \\ 
      &= \left \|\sum_{j=1}^{M}(p_j - p'_j)L_j(r) \right \|
\end{align*}
We prove (\ref{eq:inequation}) by solving the following convex optimization problem:
\begin{align*}
    &\max \left \|\sum_{j=1}^{M}(p_j - p'_j)L_j(r) \right \|, \\ &\text{subject to:}
    \begin{cases}
        \sum_{j=1}^{M}(p_j - p'_j)^2 &\leq \delta^2, \\
        \sum_{j=1}^{M}p_j - 1 &= 0.
    \end{cases}
\end{align*}
Let us define \( f(p') = \sum_{j=1}^{M}(p'_j - p_j)L_j(r) \). The problem mentioned above can then be addressed by finding the minimum and maximum values of \( f(p') \). This can be done by applying the Karush Kuhn Tucker (KKT) conditions.
Since \( f(p') \) is an affine function, the solution of the KKT conditions is the global solution of the following system of conditions:
%
\begin{align}
\hspace{-3pt}
    \left\{
    \begin{array}{l}
        \sum_{j=1}^{M}(p_j - p_j')^2 \leq \delta^2, \\[5pt]
        \sum_{j=1}^{M}p_j - 1 = 0, \\[5pt]
        u \left[ \sum_{j=1}^{M}(p_j - p_j')^2 - \delta^2 \right] = 0, \\[5pt]
         \label{2.4} L_j(r) - 2a(p'_j- p_j) + v = 0 \quad \text{for } j = \{1, \dots, M\},
    \end{array}
    \right.
\end{align}
where $u, v \in \mathbb{R} $ are Lagrange multipliers.
By summing up the two sides of Eq.~\ref{2.4}, we obtain:
\begin{align*}
    \sum_{j=1}^{M}L_j(r) +  Mv = 0   \quad \Leftrightarrow \quad v = -\frac{\sum_{j=1}^{M}L_j(r)}{M}.
\end{align*}
Denote \(\bar{L} = \frac{\sum_{i=1}^{M}L_j(r)}{M}\) and substitute the value of \(b\) into the two sides of Eq.~\ref{2.4}, we have: 
\begin{align*}
    L_j(r) - 2u(p'_j- p_j) - \bar{L} = 0  \Leftrightarrow  2u(p'_j- p_j) = L_j(r) - \bar{L}.
\end{align*}
Since \(u \neq 0\), we have:
\begin{align*}
    \begin{cases}
        p'_j- p_j &= \frac{L_j(r) - \bar{L}}{2u}, \\
        \sum_{j=1}^{M}(p'_j- p_j)^2 &= \delta^2.
    \end{cases}
\end{align*}
Solving this system, we achieve the final solution:
\begin{align*}
    \begin{cases}
        v &= -\bar{L}, \\
        4u^2 &= \frac{\sum_{j=1}^{M}(L_j(r) - \bar{L})^2}{\delta^2}, \\
        p'_j- p_j &= \frac{L_j(r) - \bar{L}}{2u}.
    \end{cases}
\end{align*}
As shown above, the solution of the KKT equations is the global solution of the optimization problem. 
Applying this to \( f(p') \), we have:
\begin{align*}
    \min f(p') &= \sum_{j=1}^{M} \frac{(L_j(r) - \bar{L})L_j(r)}{2u} \\
               &= -\delta \frac{\sum_{j=1}^{M}(L_j(r) - \bar{L})L_j(r)}{\sqrt{\sum_{j=1}^{M}(L_j(r) - \bar{L})^2}}.
\end{align*}
On the other hand, we have:
\begin{align*}
    \sum_{j=1}^{M}(L_j(r) - \bar{L})^2 
    &= \sum_{j=1}^{M}(L_j(r) - \bar{L})L_j(r).
\end{align*}
Therefore:
\begin{align}
\label{eq:min}
    \min f(p') &= -\delta \sqrt{\sum_{j=1}^{M}(L_j(r) - \bar{L})^2}.
\end{align}
Similarly, we can prove that:
\begin{align}
\label{eq:max}
    \max f(p') = \delta \sqrt{\sum_{j=1}^{M}(L_j(r) - \bar{L})^2}.
\end{align}
From \eqref{eq:min}, \eqref{eq:max}, and \( H = \left \| f(p') \right \| \), the inequality (\ref{eq:inequation}) or Theorem 1 is proven with:
\begin{equation}
    \label{bound}
    H \leq \delta \sqrt{\sum_{j=1}^{M}(L_j(r) - \bar{L})^2} = \delta Q.
\end{equation}
This result indicates that the estimation error decreases when $\delta$ is smaller and the certified accuracies of the classes are more similar.


%% file: Chapters/3-2-Methodology.tex
\begin{figure*}[t]
    \centering
    \includegraphics[width=0.88\textwidth]{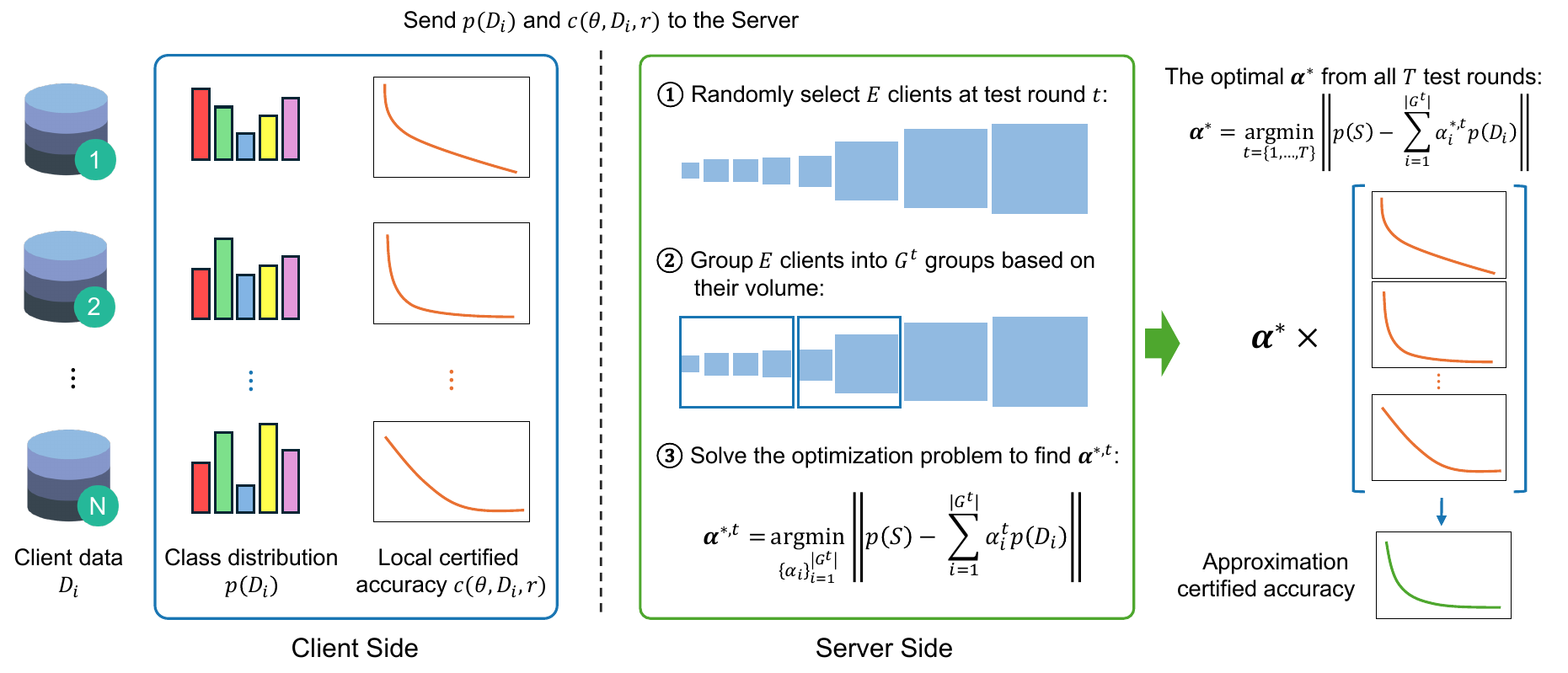}
    \caption{\textbf{Overview of FedCert.} The clients validate the accuracy of the global model using their local datasets, the server consolidates small clients into a virtual large client, and then combines all the locally validated accuracies to estimate the final certified accuracy of the global model.
    }
    \label{fig:proposed_model}
    \vspace{-0.3cm}
\end{figure*}

\subsection{Practical Algorithm}
\label{subsec:practical_algorithm}
\subsubsection{Overview}
From the theoretical analysis presented in the previous section, it becomes evident that accurately estimating the global model's accuracy necessitates addressing two critical challenges: (1) It is essential to certify the accuracy of $\theta$ on clients' local datasets \((D_{i_1}, \ldots, D_{i_m})\), which possess a sufficiently large volume. (2) It is necessary to determine a linear combination of \((p(D_{i_1}), \ldots, p(D_{i_m}))\) that closely approximates \(p(S)\).

To address the first problem, our approach involves grouping clients with smaller datasets into virtual clients with a sufficiently large amount of data. In addition, we propose an algorithm that estimates the certified accuracy of $\theta$ with respect to the \emph{virtual} dataset of each virtual client while ensuring the privacy of the clients' local data.

For the second problem, we formulate it as a convex optimization problem and solve it using an optimization solver. Specifically, our algorithm consists of three steps (Fig.~\ref{fig:proposed_model}):
\begin{itemize}
    \item \textbf{Local Accuracy Certification:} Each client \(i\) certifies the accuracy of the global model \(\theta\) using its local dataset \(D_i\).
    \item \textbf{Volume-based Client Grouping:} Clients with small datasets are grouped into virtual clients with sufficiently large volume of data. The certified accuracy of \(\theta\) on the \emph{virtual} data of each virtual client is calculated based on the certifications of the individual clients that comprise it.
    \item \textbf{Global Accuracy Certification:} The final certified accuracy of the global model is determined by combining the locally certified accuracies from the large clients (those with substantial data that are not grouped) and the accuracies computed by the the virtual clients.
\end{itemize}
In the following, we provide a detailed description of each step.


\subsubsection{Local Accuracy Certification}
\label{method:ca-determniation}
Inspired by \cite{cohen2019certified}, in our framework, each client employs the CERTIFY algorithm to certify the global model $\theta$ on its local data.  
Given an input $x$ from a client, the CERTIFY algorithm draws $n_0$ samples $\{\theta_1, \ldots, \theta_{n_0}\} \sim \mathcal{N}(x, \sigma I)$ and feeds these into the model to identify the top class $A$. 
To determine the certified radius $R$ for $x$, the algorithm draws $n$ samples $\{\theta_1, \ldots, \theta_n\} \sim \mathcal{N}(x, \sigma I)$ and estimates the lower bound of $p_A$, the proportion of class $A$ within the $L_2$ norm of radius $\sigma$. This is calculated using a one-sided $1 - \alpha$ lower confidence interval for the Binomial parameter $p$. If $\underline{p_A} > \frac{1}{2}$, CERTIFY returns the prediction $c_A$ and radius $R$; otherwise, it abstains from making a prediction, indicating that the model does not robustly trust its prediction for the input $x$.
\begin{equation*}
    R = \sigma \phi^{-1}(\underline{p_A}),
\end{equation*}
where $\phi^{-1}$ is the inverse of the standard Gaussian CDF and $\sigma$ is the noise level. The certified accuracy of client $D_i$ can be computed based on the output of CERTIFY algorithm.
Then the client sends its certified accuracy and label distribution to the central server to start the approximation phase.


    
        
        
\subsubsection{Volume-based Client Grouping}
\begin{algorithm}[t]
\caption{Grouping Algorithm}
\label{alg:grouping}
\small
\begin{algorithmic}[1]
    \State \textbf{Input:} Small clients \(\mathcal{SC}\), large clients \(\mathcal{LC}\), threshold \(\tau\)

    \State Sort \(\mathcal{SC}\) in ascending order of data size \(n_i\)
    \State Initialize \(\mathcal{V} \gets \emptyset\); \(Q \gets \text{Queue}(\mathcal{SC})\)
    \While{$Q \neq \emptyset$}
        \State Initialize virtual client \( V \gets \emptyset \)
        \While{\(n_{V} < \tau\) \textbf{and} \(Q\) is not empty}
            \State \(C \gets Q.\text{dequeue}()\); \(V \gets V \cup C\)
        \EndWhile
        
        \State Add \( V \) to \(\mathcal{V}\)
    \EndWhile
    \State \(G \gets\) \(\mathcal{LC}\) \(\cup\) \(\mathcal{V}\)
    \State \textbf{Return} \(G\)
\end{algorithmic}
\end{algorithm}
To address the problem of unreliable accuracy certification from clients with insufficient data (referred to as \textit{small clients} hereafter), we propose a grouping algorithm that merges all small clients into larger virtual clients. 
First, we identify small clients using a predefined threshold $\tau$, meaning that any client with a data size smaller than $\tau$ is classified as a small client.
The small clients are then sorted according to the size of their datasets. 
After that, small clients are incrementally grouped into virtual clients, ensuring that the total number of samples for each virtual client is no less than $\tau$. The details of the grouping process are described in Algorithm \ref{alg:grouping}.

Suppose $V$ is a virtual client made by grouping $m$ small clients $(C_{i_1}, \ldots, C_{i_m})$, we denote by $D_V$ and $p(D_V)$ the virtual dataset and virtual class distribution of $V$, and define the values of $p(D_V)$ and $c(\theta, D_V, r)$ as follows.
\begin{equation*}
    \begin{cases}
    p(D_V) &= \sum_{j=1}^{m} \frac{n_j}{n_V} p(D_j)\\[5pt]
    c(\theta, D_V, r) &= \sum_{j=1}^{m} \frac{n_j}{n_V} c(\theta,D_j, r), 
    \end{cases}
\end{equation*}
where $D_{j}$ is the dataset of $C_{i_j}$, and $n_j$ is the cardinality of $D_{j}$, $n_V$ is the sum of all $n_j$ $(j = \{1, \ldots, m\})$.
\subsubsection{Global Accuracy Certification}
\label{subsub:approximation_of_desired_distribution}
\begin{algorithm}[t]
\caption{Global Accuracy Certification}
\label{alg:approximation}
\small
\begin{algorithmic}[1]
    \For{$t = 1$ \textbf{to} $T$}
        \State Randomly select $E$ clients
        \State Group $E$ clients using Algorithm 1 to get $G^t$
        \State Solve the optimization problem to find $\boldsymbol{\alpha}^{*,t}$:
        \begin{align*}
            &\boldsymbol{\alpha}^{*,t} = \underset{\{\alpha_i\}_{i=1}^{|G^t|}}{\text{argmin}} \left\|p(S) - \sum_{i=1}^{|G^t|} \alpha_i^t p(D_i) \right\|, \\
            &\text{subject to} \quad \sum_{i=1}^{|G^t|} \alpha_i^t = 1, \quad 0 \leq \alpha_i^t \leq 1 \quad \forall i \in [1, {|G^t|}].
        \end{align*}
    \EndFor
    \State $\boldsymbol{\alpha}^{*}, G^{*} = \underset{t = \{1,\ldots,T\}}{\text{argmin}} \left\|p(S) - \sum_{i=1}^{|G^t|} \alpha^{*,t}_i p(D_i) \right\|$
    \State $c(\theta, S, r) \approx \sum_{i=1}^{|G^{*}|} \alpha_i^* c(\theta, D_i, r)$
    \State \textbf{Return}: $c(\theta, S, r)$
\end{algorithmic}
\end{algorithm}
With the locally certified accuracies in hand, the server now aggregates them to estimate the accuracy of the global model \(\theta\) on the test set \(S\). 
The central idea in this step is to find a linear combination of local class distributions (including those from large clients and the virtual clients) that best approximates \(p(S)\). 
This can be done by solving a convex optimization problem. 
Specifically, let $G$ be the set of large clients and the virtual clients obtained by Algorithm~\ref{alg:grouping}, and \( (D_1, \ldots, D_{|G|}) \) denote their datasets. 
We then use CVXPY~\cite{diamond2016cvxpy}, an optimization solver, to solve the following problem:
\begin{align*}
    &\boldsymbol{\alpha^*} = \underset{\{\alpha_i\}_{i=1}^{|G|}}{\text{argmin}} \left\|p(S) - \sum_{i=1}^{|G|} \alpha_i p(D_i) \right\|,\\
    &\sum_{i=1}^{|G|} \alpha_i = 1, \quad 0 \leq \alpha_i \leq 1, \quad \forall i \in [1, {|G|}].
\end{align*}
With the optimal values $\boldsymbol{\alpha^*}$ in hand, we estimate the certified accuracy of $\theta$ with respect to $S$ as follows:
\begin{equation}
    \label{eq:approximate}
    c(\theta, S, r) \approx \sum_{i=1}^{|G|} \alpha^{*}_i c(\theta, D_i, r).
\end{equation}
To improve the precision of the final estimated certified accuracy, we introduce an additional enhancement as follows (see Algorithm \ref{alg:approximation}). 
Rather than executing the second and third steps (i.e., volume-based client grouping and global accuracy certification) with all clients at once, we perform \( T \) iterations. 
In each iteration $t \in T$ , we randomly select only \(E\) clients $(E  \leq  N)$ to perform the second and third steps. Ultimately, we choose the result of iteration that produces the smallest error between \(p(S)\) and the linear combination of the clients' class distributions.


%% file: Chapters/4-Experiments.tex
This section evaluates the performance of the proposed FedCert method in estimating the certified accuracy of a model trained on an FL system.
We train ResNet-18~\cite{he2016deep} 
and MobileNetV2~\cite{sandler2018mobilenetv2} 
using three well-known FL settings, FedAvg~\cite{mcmahan2017communication}, FedProx~\cite{li2020federated}, and Scaffold~\cite{karimireddy2020scaffold}. 
We then perform the testing phase using the model achieved from the training phase to estimate the certified accuracy.
In the following, we first describe the setup for the experiments in Section~\ref{sec:exp_setting}. We then report and compare the performance of the FedCert with the Volume-based Weighted-sum method (\textbf{VW}) \cite{https://doi.org/10.48550/arxiv.2210.13291} on various datasets and non-IID settings in Section~\ref{sec:exp_result}. 
\textbf{VW} method aggregates the clients' certified accuracy using a weighted-sum approach, where the weights are based on the number of samples each client contributes.
For FedCert, we use ``\textbf{AP}'' to present the result of the approximation method without client grouping and ``\textbf{GA}'' to refer to the proposed method with integrated client grouping.

\subsection{Experimental Settings}
\label{sec:exp_setting}
\noindent \textbf{Datasets:} 
We use two benchmark imaging datasets frequently used in the FL~\cite{mcmahan2017communication, li2020federated} in this evaluation, i.e., CIFAR-10 and CIFAR-100. 
We split each dataset into 50,000 images for the local datasets and 10,000 images for the target test dataset $S$. 
The local datasets are distributed to clients  using different types of non-IID distributions: 
\begin{itemize}
    \item \textbf{Pareto:} The number of images of each class among clients follows a Power law distribution, formulated as \( P(X > x) = \left(\frac{x_m}{x}\right)^{\beta} \), where \( x_m \) is the scale parameter and \( \beta \) is the shape parameter.
    \item \textbf{Dirichlet:} The samples are partitioned among each client by sampling proportions from a Dirichlet distribution, formulated as \( \pi \sim \text{Dirichlet}(\beta) \), where \( \beta \) is the concentration parameter.
\end{itemize}
The local data of each client is then divided into local train and local test sets\footnote{The local test sets of a client $i$ are referred to as $D_i$ in Section~\ref{sec:mothodology}.} with an 80:20 ratio. 
Unless otherwise mentioned, the default settings are $\beta=0.1$ for Dirichlet and $\beta=3$ and $5$ for Pareto with CIFAR-10 and CIFAR-100 datasets, respectively.

\noindent \textbf{Setting for Training Phase:} 
We use Stochastic Gradient Descent (SGD) as the local optimizer, with local epochs set to $5$ and a learning rate of $0.01$. 
The clients' local data is divided among $100$ clients, with $10$ clients participating in each of the $1000$ communication rounds. 
We also set the proximal term to $0.01$ for FedProx, and the global step-size to $1.0$ for Scaffold, as suggested in the original work~\cite{li2020federated,karimireddy2020scaffold}. 
Moreover, to enhance the robustness of the global model, we implement adversarial training by adding Gaussian noise $\mathcal{N}(0,0.1)$ to each sample before feeding it into the model. 

\noindent \textbf{Setting for Testing Phase:} For a fair comparison, we use the same learned settings for testing with both \textbf{AP}, \textbf{GA}, and \textbf{VW}. 
For the approximation methods, unless otherwise mentioned, the default settings are $T=1000$ rounds, and $E=10$ (see Algorithm~\ref{alg:approximation}).
For the grouping algorithm, the sample threshold (line 5 in Algorithm~\ref{alg:grouping}) is set to $\tau=50$.


\noindent \textbf{Evaluation metrics:} 
We employ two key metrics: Root Mean Squared Error (RMSE) and Mean Absolute Percentage Error (MAPE), to compare the approximated certified accuracy with the ground truth certified accuracy: 
\begin{align*}
    \text{RMSE} &= \sqrt{\frac{1}{N}\sum_{r}(c_{approx}(\theta, S, r) - c(\theta, S, r))^2}, \\
    \text{MAPE} &= \frac{1}{N}\sum_{r} \left \| \frac{c_{approx}(\theta, S, r) - c(\theta, S, r)}{c(\theta, S, r)} \right \|,
\end{align*}

where $r \in \{0, \frac{1}{N}, \frac{2}{N}, \dots, 1\}$ (i.e., $N = 20$), and $S$ is the test dataset mentioned above. 
The ground truth certified accuracy $c(\theta, S, r)$ is calculated directly on the target dataset $S$ by the algorithm presented on \ref{method:ca-determniation}. The approximation certified accuracy $c_{approx}(\theta, S, r)$ is obtained by aggregating the certified accuracy of clients based on Eq.~\ref{eq:approximate}.

\subsection{Experimental Results}
\label{sec:exp_result}

\subsubsection{Performance of approximation methods}
\begin{table}[t]
\centering
\caption{Performance of three approximation methods for estimating certified accuracy with different FL settings. 
RMSE and MAPE show the error of the approximated certified accuracy compared to the ground truth certified accuracy.}
\label{table:diff-backbone}
\setlength\tabcolsep{4pt} 
\resizebox{\linewidth}{!}{%
\begin{tabular}{c|ll|ccc|ccc}
\toprule
 \multirowcell{2}[-2pt]{}  &
 \multirowcell{2}[-2pt]{\textbf{Dataset}} & \multirowcell{2}[-2pt]{\textbf{Client} \\ \textbf{Partition}} 
 & \multicolumn{3}{c}{\textbf{RMSE}} 
 & \multicolumn{3}{c}{\textbf{MAPE}} \\ 
 \cmidrule(lr){4-6} \cmidrule(lr){7-9} 
  & & & \textbf{AP} & \textbf{GA} & \textbf{VW} &
  \textbf{AP} & \textbf{GA} & \textbf{VW} \\ 
\midrule
\multirow{4}{*}{\rotatebox{90}{\textbf{Resnet-18}}} &
CIFAR-10 & Dirichlet 
& 0.021 & \textbf{0.014} & 0.061 & 0.059 & \textbf{0.055} & 0.192 \\ 
& CIFAR-10 & Pareto 
& 0.014 & \textbf{0.008} & 0.032 & 0.044 & \textbf{0.016} & 0.102 \\  \cmidrule(lr){2-9}
& CIFAR-100 & Dirichlet 
& 0.061 & \textbf{0.036} & 0.056 & 0.464 & \textbf{0.273} & 0.445 \\ 
& CIFAR-100 & Pareto 
& 0.019 & \textbf{0.007} & 0.052 & 0.370 & \textbf{0.187} & 1.036 \\ 
\midrule
\multirow{4}{*}{\rotatebox{90}{\textbf{Mobilenetv2}}} &
CIFAR-10 & Dirichlet 
& 0.103 & \textbf{0.050} & 0.109 & 0,285
 & \textbf{0.145} & 0.337 \\ 
& CIFAR-10 & Pareto 
& 0.034 & \textbf{0.009} & 0.062 & 0.249 & \textbf{0.048} & 0.556 \\  \cmidrule(lr){2-9}
& CIFAR-100 & Dirichlet 
& 0.003 & \textbf{0.001} & 0.006 & 0.187 & \textbf{0.039} & 0.060 \\ 
& CIFAR-100 & Pareto 
& 0.008 & \textbf{0.005} & 0.060 & 0.227 & \textbf{0.084} & 1.579 \\ 
\bottomrule
\end{tabular}%
}
\end{table}

Table~\ref{table:diff-backbone} presents the error in approximating the certified accuracy of three approximation methods with different settings of the training process. 
Overall, in most of the settings, \textbf{GA} consistently outperforms both \textbf{AP} and \textbf{VW} methods. 
Specifically, for the CIFAR-10 dataset with a Dirichlet partition, \textbf{GA} achieves the lowest RMSE of $0.014$ for ResNet-18 and $0.050$ for MobileNetV2. 
Similarly, in the Pareto partition, \textbf{GA} again shows superior performance, particularly for ResNet-18 with an RMSE of $0.007$. 
On the CIFAR-100 dataset, \textbf{GA} maintains its advantage, with the lowest RMSEs observed across Dirichlet and Pareto partitions. 
These results highlight the effectiveness of client grouping in improving the performance of FL systems across different settings.

\begin{table}[t]
    \caption{Impact of the data distribution on the performance of proposed methods (ResNet-18, CIFAR-10 dataset, FedAvg).}
    \label{table:client_partition}
    \centering
            \begin{tabular}{c|p{0.6cm}|ccc|ccc}
            \toprule
            \multirowcell{2}[-2pt]{\textbf{Client} \\ \textbf{Partition}} & \centering \multirow{2}{*}[-2pt]{\textbf{$\beta$}} & \multicolumn{3}{c}{\textbf{RMSE}} & \multicolumn{3}{c}{\textbf{MAPE}} \\
            \cmidrule(lr){3-5} \cmidrule(lr){6-8}
            & & \textbf{AP} & \textbf{GA} & \textbf{VW} & \textbf{AP} & \textbf{GA} & \textbf{VW}\\
            \midrule
            \multirow{6}{*}{\rotatebox{90}{Dirichlet}} & \centering 0.1 & 0.021 & \textbf{0.014} & 0.061 & 0.059 & \textbf{0.055} & 0.192 \\
            & \centering 0.3 & 0.046 & \textbf{0.025} & 0.122 & 0.179 & \textbf{0.073} & 0.464 \\
            & \centering 0.5 & 0.037 & \textbf{0.014} & 0.088 & 0.106 & \textbf{0.032} & 0.252 \\
            & \centering 1   & \textbf{0.053} & 0.065 & 0.142 & \textbf{0.124} & 0.181 & 0.447 \\
            & \centering 2   & \textbf{0.030} & 0.079 & 0.134 & \textbf{0.126} & 0.330 & 0.576 \\
            & \centering 3   & \textbf{0.033} & 0.053 & 0.152 & \textbf{0.077} & 0.153 & 0.475 \\
            \midrule
            \multirow{5}{*}{\rotatebox{90}{Pareto}} & \centering 2 & 0.026 & \textbf{0.011} & 0.125 & 0.113 & \textbf{0.049} & 0.552 \\
            & \centering 3 & 0.014 & \textbf{0.008} & 0.032 & 0.044 & \textbf{0.016} & 0.102 \\
            & \centering 4 & 0.021 & \textbf{0.017} & 0.024 & 0.146 & \textbf{0.110} & 0.155 \\
            & \centering 5 & 0.017 & \textbf{0.005} & 0.122 & 0.054 & \textbf{0.011} & 0.364 \\
            & \centering 6 & 0.019 & \textbf{0.005} & 0.052 & 0.038 & \textbf{0.011} & 0.112 \\
            \bottomrule
            \end{tabular}
\end{table}

\subsubsection{Impact of the non-IID degree} 
We study the robustness of our method with different degrees of non-IID. 
Specifically, for the Dirichlet distribution, we vary the concentration parameter $\beta$ from $[0.1, 0.3, 0.5, 1, 2, 3]$, where smaller $\beta$ values indicate a higher degree of non-IID. 
For the Pareto distribution, we change the scale parameter $\beta$ values from $[2, 3, 4, 5, 6]$, assessing the influence of different degrees of data imbalance among clients on the model's performance.

As shown in  Table~\ref{table:client_partition}, for the Pareto partition, \textbf{GA} consistently shows superior performance with the lowest RMSE and MAPE values in most cases. 
Notably, at $\beta = [5, 6]$, \textbf{GA} achieves the lowest RMSE ($0.005$) and MAPE ($0.011$), demonstrating its effectiveness in managing imbalanced data distributions. 
Compared to the \textbf{VW} method, both \textbf{AP} and \textbf{GA} significantly improve performance. 
For the Dirichlet partition, \textbf{GA} outperforms both \textbf{AP} and \textbf{VW} methods at $\beta = [0.1, 0.3, 0.5]$. 
Specifically, at $\beta = 0.1$, \textbf{GA} achieves the lowest RMSE, i.e., $0.014$, and MAPE, i.e., $0.055$, indicating its robustness in handling highly skewed data. 
However, with a lower degree of non-IID (e.g., $\beta = [1, 2, 3]$) and Dirichlet distribution, \textbf{AP} shows competitive performance and outperforms \textbf{GA}, suggesting that as the data distribution becomes less skewed, \textbf{AP} can maintain a high level of accuracy. 
This is an expected result because the grouping algorithm is proposed to tackle the unreliable certified accuracy for clients $i$ with a small number of samples $n_{D_i}$ (small clients). 
When the data distribution becomes less skewed in the Dirichlet distribution, the number of samples between clients is quite balanced around the threshold $\tau$. 
Grouping the data from two or more clients makes the number of samples in the newly formed group imbalanced, which impacts the approximation process.


\subsubsection{Robustness to the training algorithm}
In this experiment, we evaluate the performance of proposed methods with different FL algorithms used in the training process, e.g.,  FedAvg~\cite{mcmahan2017communication}, FedProx~\cite{li2020federated}, and Scaffold~\cite{karimireddy2020scaffold}. 
Table~\ref{table:diff-fedalg} shows the RMSE and MAPE results for ResNet-18 trained on the CIFAR-10 dataset under the Dirichlet partition with $\beta = 0.1$. 
The results indicate that the \textbf{GA} method consistently outperforms both the \textbf{AP} and \textbf{VW} methods across all metrics for both algorithms. 
Specifically, for the FedAvg algorithm, \textbf{GA} achieves the lowest RMSE (0.014) and MAPE (0.055), demonstrating its effectiveness in improving model accuracy. 
Similarly, for the FedProx algorithm, \textbf{GA} again shows superior performance with the lowest RMSE~(0.096) and MAPE~(0.386).

\subsubsection{Different desired data distributions}

\begin{table}[t]
\centering
\caption{Robustness of the proposed methods to the FL algorithm (ResNet-18, CIFAR-10 dataset, Dirichlet, $\beta=0.1$).}
\label{table:diff-fedalg}
\resizebox{\linewidth}{!}{%
\begin{tabular}{l|ccc|ccc}
\toprule
 \multirowcell{2}[-2pt]{\textbf{FL} \\\textbf{Algorithm}} &
  \multicolumn{3}{c}{\textbf{RMSE}} &
  \multicolumn{3}{c}{\textbf{MAPE}} \\ 
\cmidrule(lr){2-4} \cmidrule(lr){5-7} &
  \textbf{AP} & \textbf{GA} & \textbf{VW} &
  \textbf{AP} & \textbf{GA} & \textbf{VW} \\ 
\midrule
FedAvg~\cite{mcmahan2017communication} & 0.021 & \textbf{0.014} & 0.061 & 0.059 & \textbf{0.055} & 0.192 \\ 
FedProx~\cite{li2020federated} & 0.121 & \textbf{0.096} & 0.128 & 0.500 & \textbf{0.386} & 0.528 \\ 
Scaffold~\cite{karimireddy2020scaffold} & 0.006 & \textbf{0.005} & 0.010 & 0.014 & \textbf{0.013} & 0.034 \\ 
\bottomrule
\end{tabular}%
}
\end{table}
\begin{figure*}[t]
    \begin{minipage}{0.4\linewidth} 
    \centering
        \includegraphics[width=\linewidth]{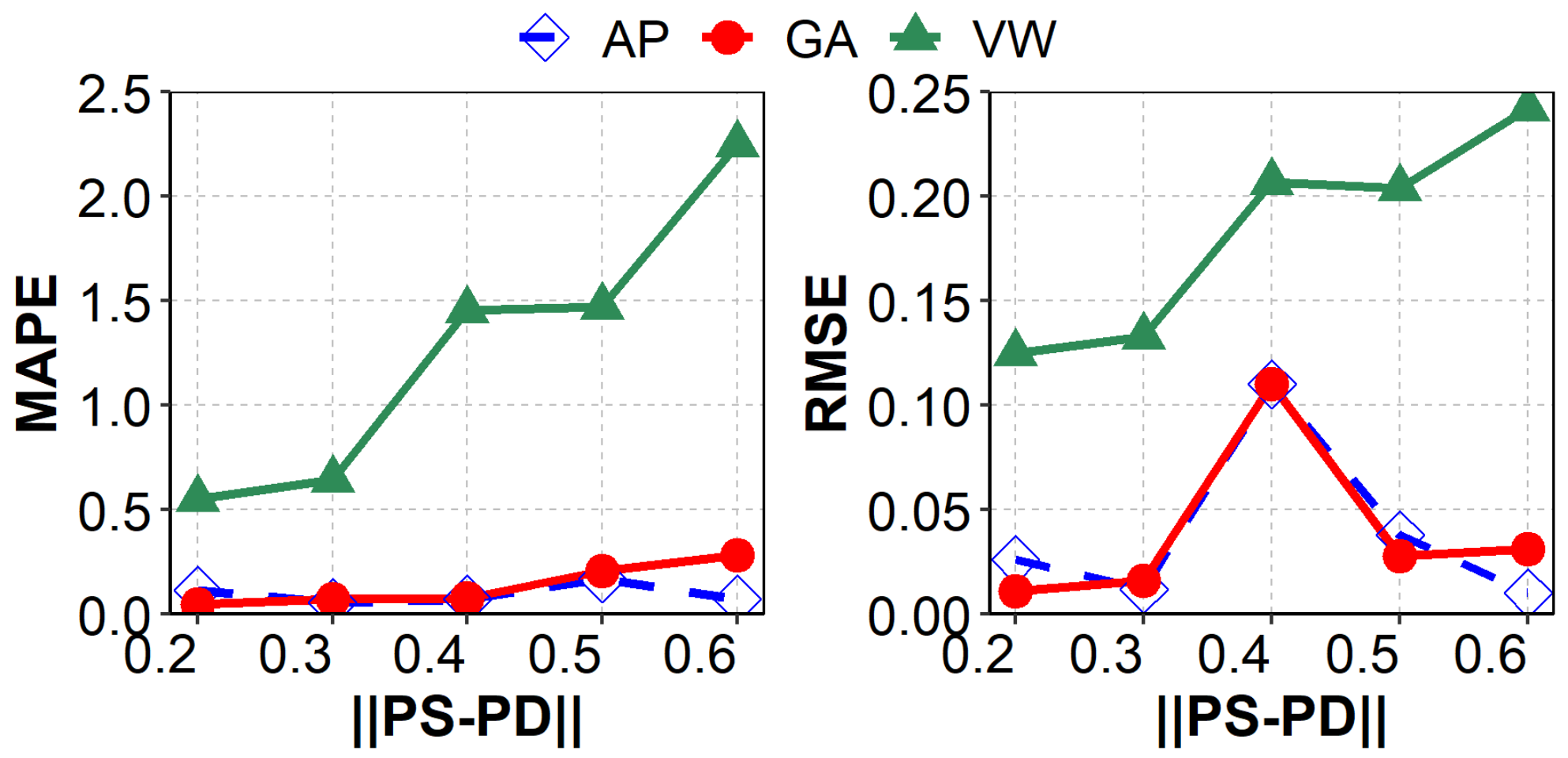}
        \caption{Performance 
        under different desired data distributions (PS) and the test sample distributions of all clients (PD). (ResNet-18, CIFAR-10 dataset, Pareto, $\beta=2$, FedAvg).}
        \label{fig:diff_distribution}
    \end{minipage}
    \hfill
    \begin{minipage}{0.305\linewidth}   
        \centering
        \captionof{table}{Impact of testing round $T$ (ResNet-18, CIFAR-10 dataset, Pareto, $\beta=2$).}
        \label{table:change_round}
        \setlength\tabcolsep{4pt} 
        \resizebox{\linewidth}{!}{%
        \begin{tabular}{c|l|ccccc}
        \toprule
        \multicolumn{2}{c|}{\textbf{T}} & 1000 & 2000 & 3000 & 5000 & 10000 \\ 
        \midrule
        \multirow{3}{*}{\rotatebox{90}{\hspace{-6pt}\textbf{RMSE}}} &
        \textbf{AP}   & 0.026 & 0.081 & 0.021 & 0.021 & 0.021 \\ \cmidrule(lr){2-7}
        & \textbf{GA} & \textbf{0.011} & \textbf{0.016} & \textbf{0.003} & \textbf{0.003} & \textbf{0.003} \\ \cmidrule(lr){2-7}
        & \textbf{VW} & 0.125 & 0.125 & 0.125 & 0.125 & 0.125\\ 
        \midrule
        \multirow{3}{*}{\rotatebox{90}{\hspace{-6pt}\textbf{MAPE}}} &
        \textbf{AP}   & 0.113 & 0.358 & 0.072 & 0.072 & 0.072\\ \cmidrule(lr){2-7}
        & \textbf{GA} & \textbf{0.049} & \textbf{0.064} & \textbf{0.009} & \textbf{0.009} & \textbf{0.009} \\ \cmidrule(lr){2-7}
        & \textbf{VW} & 0.552 & 0.552 & 0.552 & 0.552 & 0.552\\ 
        \bottomrule
        \end{tabular}
        }
    \end{minipage} 
    \hfill
    \begin{minipage}{0.253\linewidth}  
        \centering
        \captionof{table}{Impact of number of clients 
        $E$ (ResNet-18, CIFAR-10 dataset, Dirichlet, $\beta=0.5$).} 
        \label{table:change_E}
        \setlength\tabcolsep{4pt} 
        \resizebox{\linewidth}{!}{%
        \begin{tabular}{c|l|cccc} 
        \toprule
        \multicolumn{2}{c|}{\textbf{E}} & 10 & 20 & 30 & 50 \\ 
        \midrule
        \multirow{3}{*}{\rotatebox{90}{\hspace{-6pt}\textbf{RMSE}}} &
        \textbf{AP}   &  0.037 & 0.031 & 0.031 & 0.032 
        \\ \cmidrule(lr){2-6}
        & \textbf{GA} & \textbf{0.014} & \textbf{0.017} & \textbf{0.031} & \textbf{0.028} 
        \\ \cmidrule(lr){2-6}
        & \textbf{VW} & 0.088 & 0.088 & 0.088 & 0.088 
        \\\midrule
        \multirow{3}{*}{\rotatebox{90}{\hspace{-6pt}\textbf{MAPE}}} &
        \textbf{AP}   & 0.106 & 0.090 &\textbf{0.088} & 0.089 
        \\ \cmidrule(lr){2-6}
        & \textbf{GA} & \textbf{0.032} & \textbf{0.032} & 0.098 & \textbf{0.075} 
        \\ \cmidrule(lr){2-6}
        & \textbf{VW} & 0.252 & 0.252 & 0.252 & 0.252 
        \\\bottomrule
        \end{tabular}
        }
    \end{minipage}
    \vspace{-0.4cm}
\end{figure*}

We denote $\|PS - PD\|$ as the $L_2$ norm between the label distribution of test dataset $S$ and the label distribution of the union of clients' test datasets $D$. 
In this experiment, we evaluate the performance of the proposed method under different gaps between $p(S)$ and $p(D)$ to demonstrate its effectiveness across varying degrees of this disparity.
The test dataset $S$ in this experiment is created by randomly selecting samples for each class according to predefined distributions. 
The values of \( ||\text{PS} - \text{PD}|| \) used in the experiment are $[0.2, 0.3, 0.4, 0.5, 0.6]$.
The clients' local datasets are fixed for all experiments. 
To generate data for the desired distribution, the Dirichlet distribution is used to generate the vector $\text{PD} = [p_1, p_2, \ldots, p_n]$ with $n$ being the number of classes in the classification task, until achieving the desired $L_2$ norm. 
A condition of $min(\text{PD}) > 0$ is set to avoid zero probability for any class.

Fig.~\ref{fig:diff_distribution} show that the proposed methods, \textbf{AP} and \textbf{GA}, consistently achieve lower RMSE and MAPE values compared to the \textbf{VW} method. 
As the $L_2$ norm increases, the RMSE and MAPE of the \textbf{VW} method increase significantly, whereas the proposed methods show a smaller increase. 
Specifically, \textbf{AP} and \textbf{GA} demonstrate superior performance across all experimental $L_2$ norm settings, confirming their robustness and effectiveness in handling varying data distributions.

\subsection{Ablation Studies}
\noindent \textbf{Impact of number testing round $T$:} 
As shown in Table~\ref{table:change_round}, when $T$ increases, (i) \textbf{GA} still consistently outperforms both \textbf{AP} and \textbf{VW} methods and (ii) the performance of the proposed methods also improves, i.e., resulting in smaller RMSE and MAPE values. 
However, when $T=3000$, the performance becomes saturated and does not increase further. Therefore, considering trade-off between performance and computation cost, we suggest to set $T=1000$.

\noindent \textbf{Impact of number clients per round $E$:}
Table~\ref{table:change_E} shows the performance of of three approximation methods when the number of clients per round $E$ varied in $[10, 20, 30, 50]$. The results still show the superiority of \textbf{GA} over other methods in both RMSE and MAPE. Interestingly, the performance of AP and VW is not affected by $E$. In contrast, as $E$ increases, the error between approximate certified accuracy and the ground truth certified accuracy of \textbf{GA} becomes larger.
Therefore, we choose to set $E = 10$ in our experiments.

%% file: Chapters/6-Conclusion.tex
In this study, we propose FedCert, an algorithm designed to calculate certified accuracy in the FL system. 
By incorporating the client grouping algorithm and leveraging certified accuracy principles, FedCert offers a structured approach to enhance the robustness of FL models against adversarial perturbations. 
Our theoretical analysis highlights the limitations of existing aggregation methods and introduces a novel approximation approach for the desired data distribution. 
Extensive experiments on the CIFAR-10 and CIFAR-100 datasets demonstrate significant improvements in accurately evaluating the robustness of the FL system. 
These findings suggest that FedCert can be effectively applied in decentralized learning, ensuring secure and reliable FL applications. 
Future research will focus on further optimizing the algorithm and exploring its applicability to diverse datasets and FL scenarios.
The source code is available at \url{https://github.com/thanhhff/FedCert/}.